# Evaluating the Utility of GAN Generated Synthetic Tabular Data for Class Balancing and Low Resource Settings

Nagarjuna Venkata Chereddy[1,2] and Bharath Kumar Bolla[3,4]

[1] Liverpool John Moores University, Liverpool, UK
[2] UpGrad Education Private Limited, Nishuvi, Ground Floor, Worli, Mumbai - 400018, India
nagarjunareddy729@gmail.com
[3] University of Arizona
bharathbolla@arizona.edu
[4] https://orcid.org/0000-0002-4726-042X

**Abstract.** The present study aimed to address the issue of imbalanced data in classification tasks and evaluated the suitability of SMOTE, ADASYN, and GAN techniques in generating synthetic data to address the class imbalance and improve the performance of classification models in low-resource settings. The study employed the Generalised Linear Model (GLM) algorithm for class balancing experiments and the Random Forest (RF) algorithm for low-resource setting experiments to assess model performance under varying training data. The recall metric was the primary evaluation metric for all classification models. The results of the class balancing experiments showed that the GLM model trained on GAN-balanced data achieved the highest recall value. Similarly, in low-resource experiments, models trained on data enhanced with GAN-synthesized data exhibited better recall values than original data. These findings demonstrate the potential of GAN-generated synthetic data for addressing the challenge of imbalanced data in classification tasks and improving model performance in low-resource settings.

## 1    Introduction

One of the significant problems in classification tasks is a class imbalance, which occurs when one class value in the target variable is highly dominant over the other class values. It is imperative to handle the class imbalance in the target variable of classification datasets, especially in predicting rare-case scenarios. If the class imbalance is not treated, machine learning models built on the imbalanced data will suffer from unwanted bias, leading to incorrect classifications. In rare cases of classification problems, misclassification often leads to severe consequences. Several class imbalance handling techniques have been proposed over the years.

Sampling techniques like random undersampling, random oversampling, SMOTE [1], ADASYN [2], and other variants of SMOTE [3] can be used to handle the class imbalance, but they have their limitations. Due to recent advancements in deep learning, neural network models were employed to generate synthetic data. With the emergence of GANs [4], GAN-based architectures were also used in synthetic data generation. With the success of deep learning architectures in generating synthetic data on images and text, state-of-the-art architectures like GANs were extended to synthesize tabular data. From [5] and [6], it is evident that robust real-world application models need large-scale training data to yield superior performance. But large-scale data is not often readily available. In such cases, synthetic data generation techniques can curate a large-scale dataset with the same structural and statistical properties as the original data [7].

This research uses GAN-based models to create synthetic data points with similar structural and statistical properties to the original data. Synthetic data with the best qualitative properties is combined with actual data to evaluate the utility of the synthetic data in class balancing and low-resource setting experiments. The objectives of this paper are as follows:

- To evaluate the utility of synthetic data from GAN, SMOTE, and ADASYN in complementing the train data in low resource settings (classification tasks).
- To compare GAN and Non-GAN synthetic data generation techniques (SMOTE and ADASYN) on highly imbalanced data.

## 2 Literature Review

Synthetic data has gained popularity in recent years. In supervised learning, with artificial data generation techniques, researchers can generate an arbitrary number of synthetic samples with labels similar to the original data samples. The research community is increasingly using synthetic data in combination with actual data in data crunch situations to overcome the shortage of data. Researchers have conducted numerous studies on different artificial data generation techniques.

### 2.1 Synthetic data using SMOTE and ADASYN

Classification models trained on imbalanced data lead to misclassifying data points from minority classes. Over-sampling and under-sampling methods can be used to treat class imbalances. Over-sampling leads to an increase in training time and overfitting, and under-sampling leads to profound data loss. SMOTE was proposed to overcome the issue of class imbalance by creating synthetic samples [1]. SMOTE assigns equal weights to all minority class samples and overgeneralizes the minority class. Due to this drawback, multiple variants of SMOTE, like Safe Level SMOTE, SMOTE-SGA, ADASYN, and Borderline-SMOTE, were proposed [1, 3]. ADASYN uses a probability distribution function and creates more synthetic minority samples for data points that are hard to learn [2]. A combination of oversampling and undersampling methods was used to treat the class imbalance [8]. This may solve the problem to some extent. However, under-sampling leads to data loss, which is not advisable on a highly imbalanced dataset, and random oversampling leads to duplicate records that result in the overfitting of the model [1, 3].

### 2.2  Synthetic Data for Images and Text

Data augmentation is a widely used technique for improving the performance of machine learning models [9], [10], [11]. However, there are cases where augmentation can decrease model performance, particularly in detecting defects [12], [13]. To address this, recent developments in deep learning have led to the use of neural network models for generating synthetic data. For example, a parametric face generator was used to create synthetic face images, which were used for training deep neural network models [14]. In another study, an artificial data generation process was developed for generating synthetic images with synthetic backgrounds. The Faster R-CNN algorithm was trained fully on these images, achieving comparable performance to real-world data [14].

Synthetic data generation techniques have been used to overcome data shortages in specific domains. For instance, a chatbot in the Filipino language was developed using synthetic data generated using Taglog Roberta due to a data shortage in the language [15]. Similarly, synthetic data was generated from discharge summaries of mental health patients to overcome the scarcity of labeled data in the healthcare domain [16].

The use of Generative Adversarial Networks (GANs) in synthetic data generation has also been explored. Deep Convolutional GANs (DCGANs) have been proposed to generate better-quality synthetic images than conventional techniques [17]. However, using GANs for text generation is difficult due to the requirement of different data generated by the generator. To overcome this, the Medical Text GAN (mtGAN) architecture was proposed, which employs the REINFORCE algorithm to generate synthetic text data for Electronic Health Records (EHRs) [18]. In mtGAN, the discriminator's outputs serve as reward points for the generator to generate high-quality synthetic data.

### 2.3  Synthetic data for Tabular Data using Generative Models

The success of GANs in generating synthetic data on images and text has led to their use in synthesizing tabular data with the same features and statistical properties as the original data [7]. A technique for generating synthetic relational databases called CTGAN was proposed, while Synthetic Data Vault (SDV) uses Gaussian copulas to understand the interdependencies of columns and generate synthetic data [19]. Synthetic data generated using SDV has been evaluated through a crowd-sourced experiment and found no significant difference in performance compared to models trained on original datasets [20]. GAN architectures, including the Synthetic Data Vault, Data Synthesizer, and Synthpop ecosystems, have been evaluated for their utility in generating synthetic data for machine learning tasks [21, 22]. Synthetic data can also protect user privacy when sensitive data needs to be shared. However, it must be generated at a greater distance from real data to increase privacy [23]. However, the authors have not addressed the class imbalance in the datasets and have used datasets with few data points, leaving scope to explore the use of tabular synthetic data to complement training data for machine learning models in low-resource settings.

## 3 Research Methodology

As part of this research, two studies were conducted on the imbalanced insurance data [24]. The steps followed in Study 1 and Study 2 were illustrated in Fig. 1. and Fig. 2. respectively.

- Study 1: In this study, the performance of models was evaluated by training Random Forest (RF) algorithm on varying combinations of original and synthetic data.

- Study 2: Generalised Linear Model (GLM) algorithm was applied to the data balanced with synthetic data from the minority class samples.

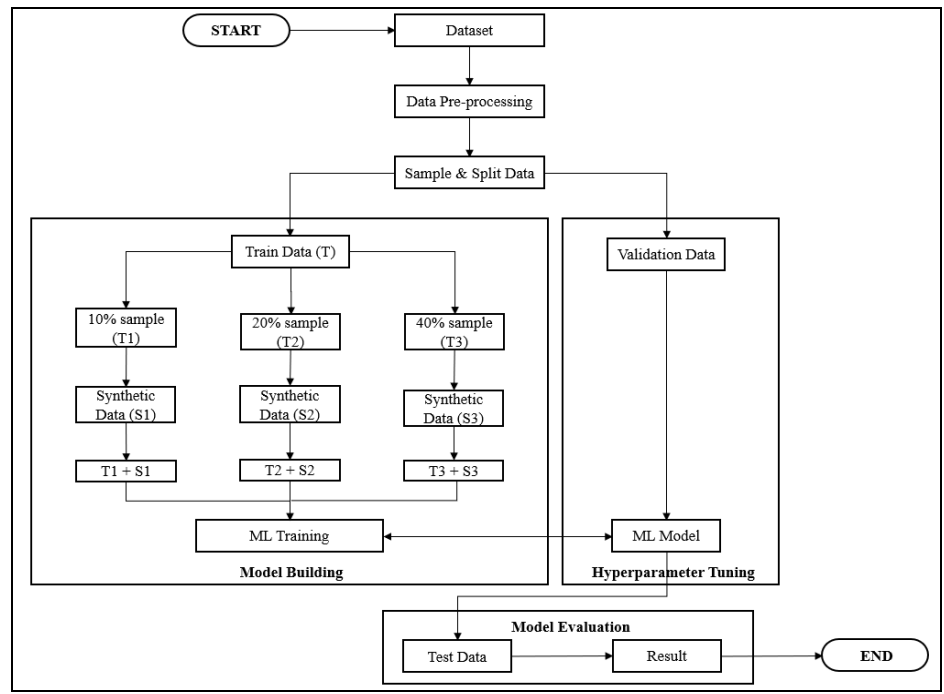

**Fig. 1.** Flow chart of the study of synthetic data in a low resource setting

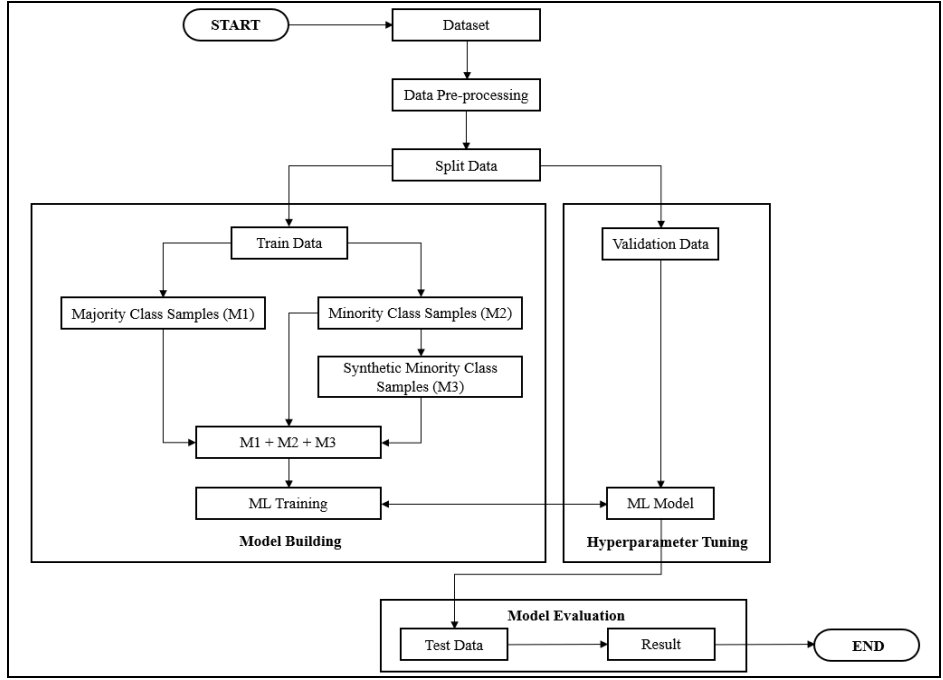

**Fig. 2.** Flow chart of the study of class balancing experiments

### 3.1 Dataset Description

As part of this research, highly imbalanced vehicle insurance data with 382154 instances was obtained from Kaggle [24]. The Response variable is the target variable with binary values 0 and 1. 0 and 1 correspond to the customer's interest in taking up the policy with the insurance company. Id is the unique identifier for each row. Age, Region_Code, Annual_Premium, Policy_Sales_Channel, and Vintage columns have numerical values. The rest of the columns are categorical/binary variables. The dataset has 62601 (16.4%) instances for which the response was flagged as one, and for the rest, the response was flagged as zero. This is a clear case of class imbalance and must be handled appropriately.

### 3.2 Data Pre-processing

Region_Code and Policy_Sales_Channel columns have 53 and 156 unique values, respectively, where each region and channel was represented with a numeric value instead of their names. So, these two columns were converted from integer to object datatype. Standard pre-processing techniques were applied to clean and prepare the data for modeling. The pre-processed data is ready for modeling, referred to as df_train_model. df_train_model dataset was used in class balancing experiments. 3 new datasets df_10_org (T1), df_20_org (T2) and df_40_org (T3) with 10%, 20% and 40% of the df_train_model were created using stratified sampling. All data points from the newly created datasets T1, T2, and T3 are used to generate S1, S2, and S3 and train classification algorithms in a low-resource setting.

### 3.3  GAN Training

In this study, we utilized two models from the Synthetic Data Vault ecosystem, namely CTGAN, and CopulaGAN, to generate synthetic data. Both models were trained for 100 epochs with a batch size of 500 and a learning rate of 0.0002 for both the generator and discriminator. The discriminator steps of both models were set to 5. Both GAN models were trained on df_10_org and synthetic data of the same size as df_10_org was generated from each model. The synthetic data generated from each model was used to enrich the df_10_org, df_20_org, and df_40_org datasets by synthesizing 26750 (S1), 53501(S2), and 107002 (S3) records, respectively. Additionally, 223686 (M3) minority class records were synthesized to balance the imbalanced df_train_model dataset. The synthetic data generated by the best-performing GAN was selected for experiments using KSTest and Logistic Detection Score metrics.

### 3.4  Data Modelling

The data used in class balancing and low resource setting experiments were split into a 70:30 ratio using stratified train-test split, and continuous variables were rescaled using StandardScaler. In order to balance and improve the rescaled train data, GAN, SMOTE, and ADASYN were utilized. For class balancing, four versions of data were available for modeling, including the original imbalanced data and data balanced with each of the SMOTE, ADASYN, and GAN techniques. Twelve versions of data were available for low resource settings, including the original data sets and the enriched data sets generated using GAN, SMOTE, and ADASYN. Both experiments used the GLM and RF algorithms to build classification models.

In the class balancing experiment, the GLM algorithm was used, as it provides coefficients for all input variables. GLM models were trained on all four versions of data using Recursive Feature Elimination (RFE) and manual feature elimination techniques to remove insignificant independent variables. The top fifteen variables returned by RFE were used to build an initial model, and manual feature elimination was performed using p-values and VIF values.

In the low resource setting experiment, Classification models were built using the RF algorithm, and the data were transformed using Principal Component Analysis (PCA); for data enriched with synthetic data from GAN, seventeen principal components were used to transform the data, while fifteen principal components were used for other data versions. PCA-transformed training data was fed to the RF algorithm for training, and hyperparameter tuning was performed using GridSearchCV on n_estimators, max_depth, min_samples_leaf, and max_leaf_nodes parameters. The roc_auc metric was used to select the best-performing model, considered the final model.

The process of model evaluation and predictions on test data were consistent for both GLM and RF models. Using the final model, the odds of test data points that the models had not seen were predicted, and labels were assigned using the optimal threshold value. Accuracy, recall, AUC, specificity, and G-Mean metrics were calculated using the labels of unseen test data, and recall was used as the primary metric for evaluation.

## 4    Results

In this section, the results obtained from the models are discussed. 4.1 sub-section discusses the qualitative assessment of synthetic data generated from GANs. The two subsequent sub-sections will discuss the results obtained from class balancing and low resource setting experiments.

### 4.1    Assessment of GAN synthetic data quality

The outcomes of the assessment of synthetic data generated by GAN are presented in Table 1. CopulaGAN synthetic data demonstrated slightly better performance than CTGAN synthetic data in KSTest, with scores of 0.99 and 0.98, respectively. The detection scores of synthetic data from CTGAN and CopulaGAN were 0.76 and 0.82, respectively. CopulaGAN required 280 minutes for the training phase and only 6 minutes and 11 seconds to produce 410930 (26750 + 53501 + 107002 + 223686) data points. CopulaGAN needed less time for training and sampling than CTGAN. Despite CTGAN synthetic data displaying somewhat better results in the detection matrix, CopulGAN synthetic data demonstrated better performance than CTGAN synthetic data in all three parameters. Therefore, the synthetic data created by CopulaGAN was employed for the class balancing and low resource setting experiments.

**Table 1.** GAN Results

|  | KSTest | Logistic Detection score | Training time (mins) | Sampling Time |
|---|---|---|---|---|
| **CTGAN** | 0.98 | 0.76 | 301 | 6min 58s |
| **CopulaGAN** | 0.99 | 0.82 | 280 | 6min 11s |

### 4.2    Results on the utility of synthetic data in Class Balancing

The findings of the class balancing experiments are presented in Figure 3. The threshold values used to calculate the reported metrics are listed in Section 4. The results indicate that the synthetic data generated by GAN yielded the highest recall value (0.84), while the data balanced with ADASYN and GAN synthetic data resulted in better performance on the G-Mean metric (0.82). All versions of the data had the same AUC score (0.88). The accuracy (0.82) and specificity (0.82) values obtained from the imbalanced and SMOTE-balanced data were better than those obtained from the data balanced with ADASYN and GAN synthetic data. The balanced data showed similar performance to the imbalanced data, but the real impact of balancing with synthetic data is reflected in the threshold values. The threshold values of the balanced data were at least twice as large as those of the imbalanced data, indicating that models trained on balanced data can make predictions with greater accuracy than those trained on imbalanced data..

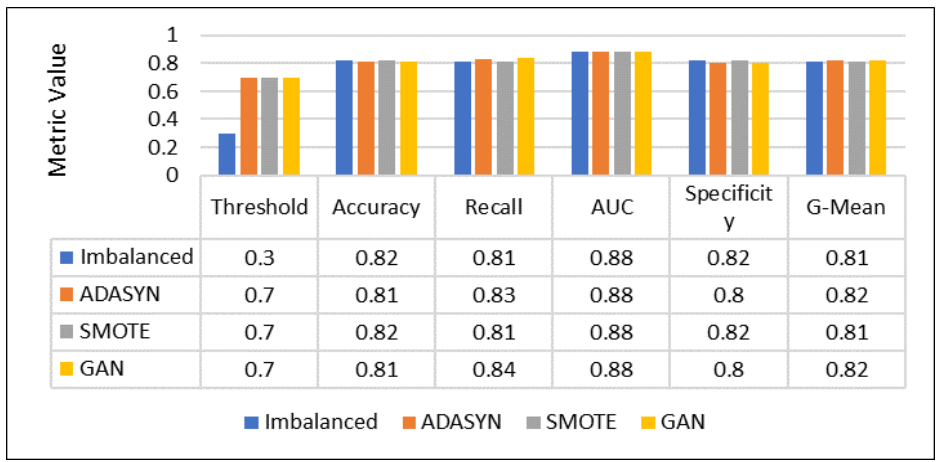

**Fig. 3.** Results of different balancing techniques

### 4.3  Low Resource Setting – Random Forest Results

The figures (Figs. 4, 5, and 6) depict the results of Random Forest on different variants of the datasets df_10_org, df_20_org, and df_40_org. The "Base" column represents the original data, while the Base + SMOTE, Base + ADASYN, and Base + GAN columns refer to the data enriched with synthetic data from SMOTE, ADASYN, and GAN, respectively. According to Fig. 4, ADASYN enhanced data has the highest accuracy (0.833), SMOTE enriched data has the highest AUC score (0.889), and GAN enriched data has the highest G-Mean (0.821). The models trained on the ADASYN and GAN enhanced data have yielded superior specificity (0.862) and recall (0.841) values, respectively. Fig. 5 shows that ADASYN enriched data has the highest accuracy (0.836), SMOTE and GAN enriched data have the highest AUC score (0.89), and GAN enriched data has the highest G-Mean (0.825). The models trained on ADASYN and GAN enriched data have yielded superior specificity (0.873) and recall (0.845) values, respectively. Fig. 6 indicates that ADASYN enhanced data has the highest accuracy (0.836) and specificity (0.867), while the original 40% of data and SMOTE and GAN enriched data have the highest AUC score (0.891). The models trained on GAN enhanced data have yielded superior recall values (0.847).

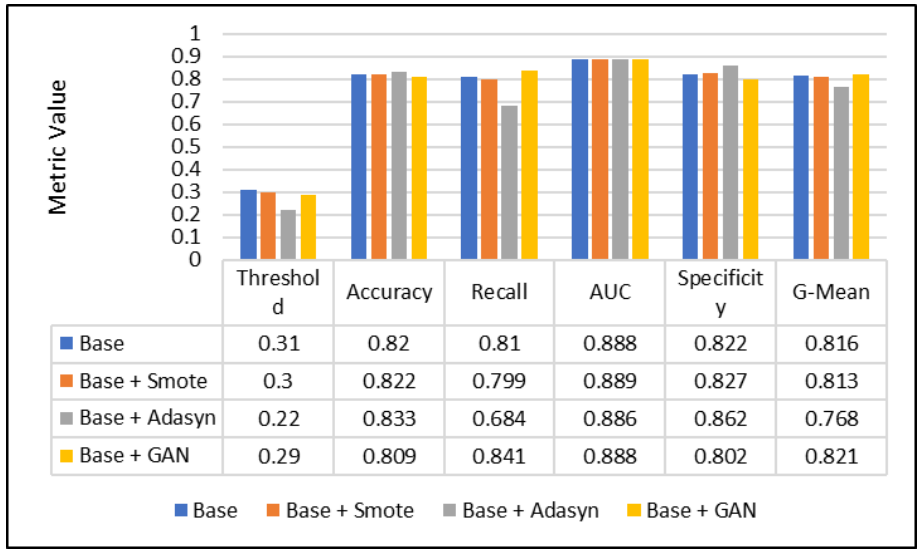

**Fig. 4.** Random Forest results with 10% of the data

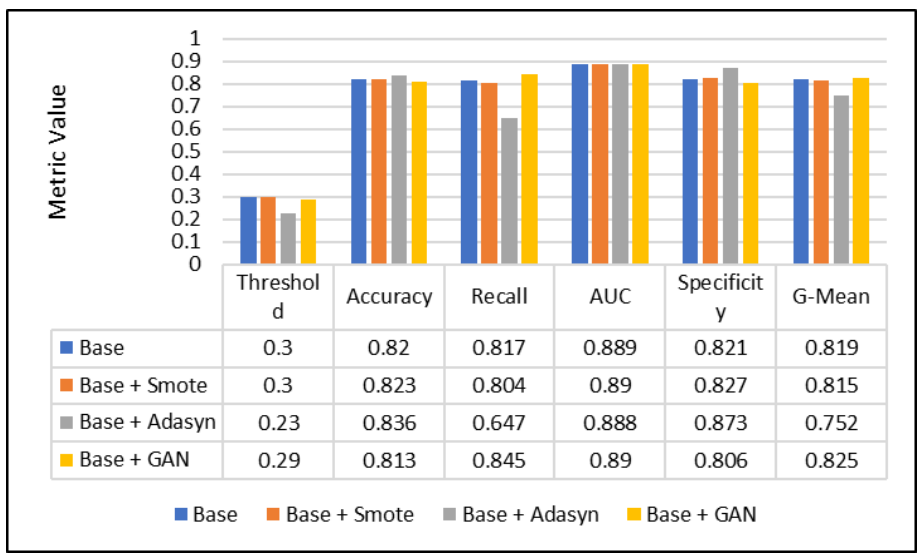

**Fig. 5.** Random Forest results with 20% of the data

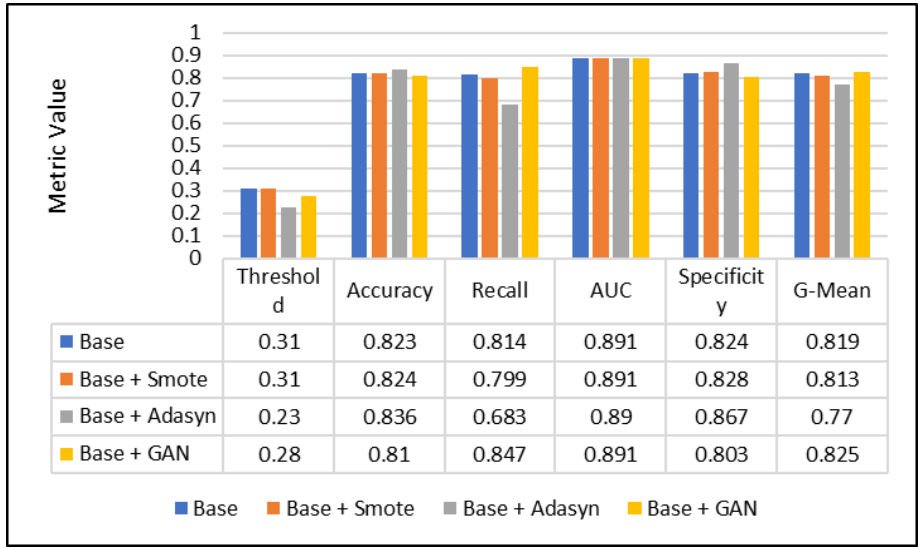

**Fig. 6.** Random Forest results with 40% of the data.

The results indicate that using GAN synthetic data can improve recall values significantly compared to using the original data. For instance, the recall value (0.841) obtained from the model trained on 10% of data enriched with GAN synthetic data is significantly higher than the recall value (0.817) obtained from the original 20% of data. This pattern is also observed when comparing the results in Fig. 5 and Fig. 6. Additionally, models trained on data enriched with SMOTE consistently yield higher AUC values, while those trained on data enriched with ADASYN yield higher accuracy and specificity values (Figs 4,5 and 6). Finally, the models trained on data enriched with GAN consistently yield higher recall and G-mean values than other data versions.

## 5     Conclusion

This study is the first to investigate the use of synthetic data in a low-resource setting. Our research establishes the effectiveness of synthetic data in building classification models for highly imbalanced data in low-resource settings. Specifically, this study is the first to explore using GAN synthetic data in low-resource settings. The models trained on balanced data using GAN have resulted in the highest recall at a threshold of 0.7. The random forest models trained on data enriched with GAN synthetic data in low-resource settings have consistently produced superior recall values. The better recall values observed in the models trained on data enriched with GAN synthetic data can be explained by the fact that GAN has effectively learned the characteristics of minority class samples, thus generating high-quality synthetic data from such samples. This finding indicates that synthetic data from GAN can effectively predict rare case scenarios, such as fraudulent transactions, loan defaulters, and rare diseases like HIV AIDS, urinary tract infections, and dengue, among others. Our work generalizes the power of Generative models to solve class imbalance problems in low-resource settings, demonstrating the potential of synthetic data generated by GAN in improving classification models' performance on highly imbalanced datasets.